\documentclass[doublecolumn,oneside,letter]{IEEEconf}
\usepackage{psfrag}
\usepackage{amsmath}
\usepackage{amssymb}
\usepackage{graphicx}
\usepackage{subfigure}
\usepackage{verbatim}
\usepackage[thinlines,thiklines]{easybmat}
\usepackage{latexsym}
\usepackage[dvipsnames]{xcolor}
\usepackage{cite}
\usepackage{bm}
\usepackage{stmaryrd}
\usepackage{multirow}
\usepackage{textcomp}
\usepackage[ruled]{algorithm2e}
\usepackage{bookmark}

\def\s{\mathop{\rm s}\nolimits}
\def\c{\mathop{\rm c}\nolimits}
\newcommand{\bs}[1]{\ensuremath{{\bm{#1}}}}

\def\rank{\mathop{\rm rank}\nolimits}

\newtheorem{lem}{Lemma}

\def\atan2{\mathrm{atan2}}
\newcommand\norm[1]{\left\lVert#1\right\rVert}

\begin{document}

\title{\bf \Large Cascaded Nonlinear Control Design for Highly Underactuated Balance Robots\thanks{This work was supported in part by the US NSF under award CNS-1932370.}}

\author{Feng Han and Jingang Yi\thanks{F. Han and J. Yi are with the Department of Mechanical and Aerospace Engineering, Rutgers University, Piscataway, NJ 08854 USA (e-mail: {fh233@scarletmail.rutgers.edu}; {jgyi@rutgers.edu}).}}

\maketitle

\thispagestyle{empty}
\pagestyle{empty}
\begin{abstract}
This paper presents a nonlinear control design for highly underactuated balance robots, which possess more numbers of unactuated degree-of-freedom (DOF) than actuated ones. To address the challenge of simultaneously trajectory tracking of actuated coordinates and balancing of unactuated coordinates, the proposed control converts a robot dynamics into a series of cascaded subsystems and each of them is considered virtually actuated. To achieve the control goal, we sequentially design and update the virtual and actual control inputs to incorporate the balance task such that the unactuated coordinates are balanced to their instantaneous equilibrium. The closed-loop dynamics are shown to be stable and the tracking errors exponentially converge towards a neighborhood near the origin. The simulation results demonstrate the effectiveness of the proposed control design by using a triple-inverted pendulum cart system.
\end{abstract}

\section{Introduction}
\label{Sec_Introduction}

Underactuated robots have less number of control inputs than that of the degree-of-freedom (DOF). Highly underactuated balance robots possess more numbers of unactuated DOFs than actuated ones. Control design for underactuated balance robots faces the design challenge of limited control actuation for simultaneously trajectory tracking and platform balance. Most existing works focus on underactuated balance systems with more actuated coordinates than unactuated ones. For instance, cart-pole system has one input with 2-DOF~\cite{Karg2020ToC, HanRAL2021}, five-link bipedal walker robot has six inputs with 7-DOF~\cite{Maggiore2013Virtual, Wester2003}, autonomous bicycle robot has two inputs with 3-DOF~\cite{Han2022Tmech, ChenIJRR2015}. There are various well-developed control frameworks for those systems including the external and internal convertible form-based control (i.e., EIC-based control)~\cite{GetzPhD}, orbital stabilization~\cite{KANT2020Orbital}, energy-shaping based control~\cite{Fantoni2000}, etc. Both the model-based control and machine learning-based control approaches are extensively studied~\cite{HanRAL2021, ChenTRO2022}. However, for highly underactuated balance robots, such as a triple passive inverted pendulum on a controlled cart (i.e., one input with 4-DOF), those control approaches might not work properly. For instance, it remains an open problem for the periodical orbit stabilization design to guarantee multiple unactuated coordinates.

For highly underactuated balance robots with more unactuated than actuated coordinates, the inherently unstable property and coupled dynamics between them impose great challenges in control system design~\cite{Ben2020Triple, FURUT1984IJC,sieber2003bifurcation}. With highly limited control actuation available, there exist great task conflicts. To reduce the design complexity, most of the existing works focus on stabilization control. Linearization of nonlinearly system and pole placement/LQR (linear quadratic regulation) techniques are popular methods~\cite{Medrano1997, Ian2018master, FURUT1984IJC, Niemann2005CEP}. The research work in~\cite{Medrano1997} presented an LQR-based robust control for a triple-invented pendulum cart system and a fault tolerant control was proposed for a double-inverted pendulum cart system using a linearized model~\cite{Niemann2005CEP}. In~\cite{Jahn2021AUTO}, the authors enhanced the inversion-based approach (e.g.,~\cite{Graichen2005CDC}) towards the stabilization of a periodic orbit of a multi-link triple pendulum on a cart. To this end, A two-point boundary value problem was formulated to obtain a nominal trajectory and control design via a linear-quadratic-Gaussian controller. However, simultaneously control of trajectory tracking and platform balance remain a challenge for highly underactuated balance robots.

Among the aforementioned control methods, the EIC-based control has been demonstrated as an effective approach for underactuated balance robots. The EIC-based control has been applied to underactuated balance robots that have more numbers of actuated than unactuated DOFs, including inverted pendulum~\cite{HanRAL2021}, autonomous bikebot~\cite{Han2022Tmech,ChenTRO2022}, and aggressive vehicle under ski-stunt maneuvers~\cite{HanAIM2022}. The unstable, unactuated subsystem is balanced onto a balance equilibrium manifold (BEM) and trajectory tracking and platform balance control are achieved simultaneously. However, the EIC-based control has not been designed for highly underactuated balance robots. In~\cite{Han2024ICRA}, we show that some of the unactuated coordinates were not able to display the designed dynamics, which resulted in unstable motion. Given its attractive feature, the EIC-based control can be potentially revised or redesigned for highly underactuated balance robots.

The main feature of the EIC-based control is to embed the balance control into the trajectory tracking design. The target profile of the unactuated subsystem is associated with the actuated subsystem motion. The motion of the actuated subsystem can be viewed as a control input to drive the unactuated subsystem to its BEM. Inspired by such an observation, we propose a cascaded EIC form (i.e., CEIC) that reformulates the original highly underactuated balance system to a series of cascaded subsystems, which are virtually actuated by their interactions. Associated with each two-subsystem is one coordinate, which accounts for the coupling and also serves as a virtual control input. We sequentially estimate and obtain the BEM and then update the control input of the subsystems sequentially. Each subsystem has been shown under active control design. Trajectory tracking and balance control can be achieved. We illustrate and demonstrate the CIEC-based control through an example of a triple-inverted pendulum on a cart. The main contribution of this work is the proposed new cascaded control framework for highly underactuated balance robots. We also for the first time reveal the controllable condition of the highly underactuated balance robots.

The rest of the paper is outlined as follows. Section~\ref{Sec_Model} presents the dynamics and EIC-based control design. In section~\ref{Sec_EIC}, we propose the cascaded EIC design. The CEIC-based control is presented in~\ref{Sec_Control}. We present the simulation results in~\ref{Sec_Result}. Finally, Section~\ref{Sec_Conclusion} discusses the concluding remarks.

\section{Highly Underactuated Balance Robots}
\label{Sec_Model}

In this section, we present the dynamics and the EIC-based control design for underacted balance robots.

\subsection{System Dynamics}

Let the generalized coordinates of underactuated balance robots be $\bm q=[q_1\cdots q_{n+m}]^T \in \mathbb{R}^{n+m}$, $n,m \in \mathbb{N}$. We partition $\bm q$ into $\bm q=[\bm q_a^T\; \bm q_u^T]^T$ with actuated coordinate $\bm q_a \in \mathbb{R}^{n}$ and unactuated $\bm q_u \in \mathbb{R}^m$. The robot dynamics for actuated and unactuated subsystems can be written into~\cite{Han2023ICRA}
\begin{subequations}
\label{Eq_Sub_Dynamics}
\begin{align}
    \mathcal S_a&: \bm D_{a a} \ddot{\bm q}_{a}+\bm D_{a u} \ddot{\bm q}_{u}+\bm C_{a} \dot{\bm q}+\bm G_{a}=\bm u,\label{Eq_Actuated}\\
     \mathcal S_u&:\bm D_{u a} \ddot{\bm q}_{a}+\bm D_{u u} \ddot{\bm q}_{u}+\bm C_{u} \dot{\bm q}+\bm G_{u}=\bm 0,\label{Eq_unactuated}
\end{align}
\end{subequations}
where $\bm{D}(\bm{q})$, $\bm{C}(\bm{q},\dot{\bm{q}})$ and $\bm{G}(\bm q)$ are the inertia, Coriolis and gravity matrices, respectively. The subscripts $aa$ ($uu$) and $ua$ and $au$ indicate the variables related to the actuated (unactuated) coordinates and coupling effects, respectively. For the convenience of representation, the dependence of matrices $\bs{D}$, $\bm C$, and $\bm G$ on $\bm q$ and $\dot{\bm q}$ is dropped. We denote $\bm H_a = \bm C_{a} \dot{\bm q}+\bm G_{a}$ and $\bm H_u=\bm C_{u} \dot{\bm q}+\bm G_{u}$.

In general, robot dynamics of unactuated subsystem $\mathcal S_u$ in~\eqref{Eq_unactuated} is intrinsically unstable. Most of the previous work focus on $\mathcal{S}=\{\mathcal S_a, \mathcal S_u\}$ with the property $n\ge m$, that is, more actuated DOFs than that of unactuated. In this work, we consider highly underactuated balance robots, i.e., $n < m$. With far less control actuation, it becomes a challenging problem for simultaneously trajectory tracking and platform balance control design~\cite{Jahn2021AUTO}.

\subsection{EIC-Based Tracking Control}

We first present the EIC-based control and discuss its limitations for highly underactuated balance robot control. Given the desired trajectory of actuated coordinates $\bm q_a^d$, the goal of the robot control is to achieve that the actuated subsystem $\mathcal S_a$ follows $\bm q_a^d$ and the unactuated, unstable subsystem $\mathcal S_u$ is balanced around unstable equilibrium, denoted by $\bm q_u^e$. Note that the unstable equilibrium $\bm q_u^e$ depends on the tracking performance of $\mathcal S_a$ and its profile needs to be estimated in real-time.

Given $\bm q_a^d$, we temporarily neglect the dynamics of $\mathcal S_u$ and the control input for $\mathcal S_a$ is designed using the feedback linearization as
\begin{align}
\label{Eq_Actuated_Control}
\bm u_a^\mathrm{ext}= \bm{D}_{a a} \bm v_a^\mathrm{ext}+\bm{D}_{au} \ddot{q}_{u}+\bm H_{a},
\end{align}
where $\bm v_a^\mathrm{ext} = \ddot{\bm q}_a^d - \bm k_{p1}\bm e_a- \bm k_{d1}\dot{\bm e}_a,
$ is an auxiliary control design. $\bm e_a = \bm q_a-\bm q_a^d$ is the tracking error and $\bm k_{p1}, \bm k_{d1}$ are control gains.

The $\bm q_u$ coordinate should be stabilized onto the BEM. Given the control input $\bm u_a^\mathrm{ext}$, the BEM is defined as instantaneous equilibrium in terms of  $\bm q_u$ as
\begin{equation}
\label{Eq_BEM}
\mathcal{E}=\left\{\bm q_u^e:  \bm \Gamma(\bm q_u^e;\bm v_a^\mathrm{ext})=\bm 0, \dot{\bm q}_u=\ddot{\bm q}_u=\bm 0\right\},
\end{equation}
where $\bm \Gamma(\bm q_u;\bm v_a^\mathrm{ext})=\bm{D}_{uu} \ddot{\bm q}_u + \bm D_{ua} \bm v_a^\mathrm{ext}+ \bm H_{u}$. The equilibrium $\bm q_u^e$ is obtained by inverting $ \bm \Gamma(\bm q_u;\bm v_a^\mathrm{ext})\big\vert_{ \dot{\bm q}_u=\ddot{\bm q}_u=\bm 0}=\bm 0$. Using the BEM $\bm q_u^e \in \mathcal{E}$ as target reference for $\mathcal S_{u}$, we redesign the $\ddot{\bm q}_a$ profile such that under $\ddot{\bm q}_a$, $\bm q_u \rightarrow \bm q_u^e$. The control is updated by incorporating the $\mathcal S_{u}$ dynamics as
\begin{align}
\label{Eq_Va_Int}
    \bm v_{a}^\mathrm{int}=-\bm{D}_{u a}^{+}(\bm H_{u}+\bm D_{uu} \bm v_u^\mathrm{int}),
\end{align}
where $\bm v_u^\mathrm{int} = \ddot {\bm q}_u^e -\bm k_{p2}\bm e_u - \bm k_{d2}\bm e_u$. $\bm{D}_{u a}^{+}=(\bm{D}_{u a}^T\bm{D}_{u a})^{-1}\bm{D}_{u a}^T$ is the generalized inverse of $\bm{D}_{u a}$.  $\bm e_u= \bm q_u- \bm q_u^e$ and $\bm k_{p2}, \bm k_{p2}$ are control gains. With the design~\eqref{Eq_Va_Int}, the final control becomes
\begin{equation}\label{Eq_Ua_Int}
  \bm u_a^\mathrm{int} =  \bm{D}_{aa} \bm v_a^\mathrm{int}+\bm{D}_{a u} \ddot{\bm q}_{u}+\bm H_{a}.
\end{equation}
The above sequentially designed control, known as EIC-based control, aims to achieve tracking of $\mathcal S_a$ and balance of $\mathcal S_u$, simultaneously\cite{GetzPhD}.

Inserting the updated control design $\bm u_a^\mathrm{int}$ into the system dynamics $\mathcal S$, we obtain
\begin{align}
\label{Eq_Sa_Ua_Int}
\ddot{\bm q}_u&=-\bm D_{uu}^{-1}(\bm D_{ua} \ddot{\bm q}_a + \bm H_u)\nonumber\\
&=-\bm D_{uu}^{-1}\left[-\bm D_{ua}  \bm{D}_{u a}^{+}(\bm H_{u}+ \bm D_{uu} \bm v_u^\mathrm{int})+ \bm H_u\right].
\end{align}
Since $\bm D_{ua}\in \mathbb{R}^{m\times n}$ and $n < m$, we have $\bm D_{ua}  \bm{D}_{u a}^{+}\in\mathbb{R}^{m\times m}$ and $\rank(\bm D_{ua} \bm{D}_{u a}^{+})=n < m$. Therefore, part of the control effect design of $\bm v_u^\mathrm{int}$ would not appear and the nonlinearity term $\bm H_u$ cannot be fully canceled at all dimensions. The unactuated subsystem $\mathcal S_u$ does not approach $\mathcal E$ in $\mathbb{R}^{m}$ as designed and the balance would not be guaranteed for highly underactuated balance robot.

\section{Cascaded EIC Form For Highly Underactuated System}
\label{Sec_EIC}

The enhanced EIC-based control has been successfully demonstrated for underactuated balance robots with $n\ge m$~\cite{Han2024ICRA}. If the system $\mathcal S$ with $n < m$ can be transferred virtually to a series of subsystems with more actuated coordinates, we can still achieve guaranteed performance. We note that $\ddot{\bm q}_a$ is used as a virtual control input when incorporating the balance control $\bm v_u^\mathrm{int}$ into control design (see~\eqref{Eq_Va_Int}). However, the $\mathcal S_u$ dynamics with respect to $\ddot{\bm q}_a$  is another underactuated system with $m$ coordinates. For such an underactuated subsystem, we can perform the EIC-based control again to $\mathcal S_u$. Following such an inspiration, in the following, we formally present our design.

The $\mathcal S_a$ dynamics under the control $\bm u$ can be solved as
\begin{equation}\label{Eq1}
 \ddot{\bm q}_a=\bm D_{a a}^{-1}\left(\bm u-\bm D_{a u} \ddot{\bm q}_u-\bm H_a\right).
\end{equation}
Substituting~\eqref{Eq1} into $\mathcal S_u$ dynamics yields
\begin{equation}\label{Eq_Su_2}
\mathcal S^1: \bm D^{(1)} \ddot{\bm q}^{(1)}+ \bm H^{(1)}= \bm B^{(1)} \bm u,
\end{equation}
where $\bm q^{(1)} =\bm q_{u}$ and
$\bm D^{(1)} =\bm D_{u u}-\bm D_{u a} \bm D_{a a}^{-1}\bm D_{a u}$, $\bm H^{(1)}=\bm H_u-\bm D_{u a} \bm D_{a a}^{-1} \bm H_a$, $\bm B^{(1)}=-\bm D_{u a} \bm D_{a a}^{-1}$. We note $\bm D_{ua}\in \mathbb{R}^{m\times n}$ and $ \bm B^{(1)}\in \mathbb{R}^{m\times n}$. Equation~\eqref{Eq_Su_2} represents another underactuated balance system with $m$ generalized coordinates and $n$ control inputs.

We partition the $\bm q^{(1)}$ coordinates into two parts as
\begin{equation*}
\bm q^{(1)}=\left[ (\bm q_a^{(1)})^T~(\bm q_u^{(1)})^T
\right]^T,
\end{equation*}
where $\bm q_a^{(1)}$ denotes the first $n$ unactuated coordinates, such that $\dim(\bm q_a^{(1)})=n$, $\dim(\bm q_u^{(1)})=m-n$.  Then we rewrite the $\mathcal{S}^1$ dynamics
\begin{subequations}\label{Eq_Su_2_sub}
\begin{align}
 \mathcal S_{a}^1: \bm D_{aa}^{(1)} \ddot{\bm q}_{a}^{(1)} + \bm D_{au}^{(1)} \ddot{\bm q}_{u}^{(1)} + \bm H_a^{(1)}= \bm B_a^{(1)} \bm u,\\
 \mathcal S_{u}^1: \bm D_{ua}^{(1)} \ddot{\bm q}_{a}^{(1)} + \bm D_{uu}^{(1)} \ddot{\bm q}_{u}^{(1)} + \bm H_u^{(1)}= \bm  B_u^{(1)} \bm u,
\end{align}
\end{subequations}
where matrix $\bm D^{(1)}$, $\bm H^{(1)}$ and $\bm B^{(1)}$ are block matrixes in proper order. Clearly,~\eqref{Eq_Su_2_sub} is in the form of an underactuated robot model, similar to~\eqref{Eq_Sub_Dynamics}. We note that the input matrix $\bm B^{(1)}$ in $\mathcal S^1$ is no longer a constant. Namely, the selection of $\dim(\bm u)=n$ generalized coordinates as the actuated ones out of $\bm q^{(1)}$ is arbitrary, as long as $\rank(\bm B_a^{(1)})=n$.

From $\mathcal S^1_a$ dynamics we can solve the $\bm q_{a}^{(1)}$ dynamics as
$\ddot{\bm q}_a^{(1)}=\bm D_{aa}^{(1)}\left(\bm B_a^{(1)} \bm u- \bm D_{au}^{(1)} \ddot{\bm q}_{u}^{(1)}-\bm H_a^{(1)}\right)$. Inserting  $\ddot{\bm q}_{a}^{(1)}$ into $\mathcal S_{u}^1$, we obtain
\begin{equation*}
\mathcal{S}^2: \bm D^{(2)} \ddot{\bm q}^{(2)}+\bm H^{(2)}=\bm B^{(2)} \bm u,
\end{equation*}
where $\bm q^{(2)}={\bm q}_{u}^{(1)}$ and $\bm D^{(2)} = \bm D_{uu}^{(1)}-\bm D_{ua}^{(1)} \left(\bm D_{aa}^{(1)}\right)^{-1} \bm D_{au}^{(1)}$, $ \bm H^{(2)} = \bm H_u^{(1)}-\bm D_{ua}^{(1)} \left(\bm D_{aa}^{(1)}\right)^{-1}\bm H_a^{(1)}$, and $\bm B^{(2)} = \bm B_u^{(1)}-\bm D_{ua}^{(1)} \left(\bm D_{aa}^{(1)}\right)^{-1}\bm B_a^{(1)}$. If $\dim(\bm q_{u}^{(2)})>\dim(\bm u)$, $\bm B^{(2)} \in \mathbb{R}^{(m-n)\times n}$ and $\mathcal S^2$ is also an underactuated balance system. We can continue to perform such a transformation. We assume that there are in total $k$ actuated subsystems (each contains $n$ coordinates) and $(k+1)$-th subsystem is fully actuated (contain last $z$ coordinates, i.e., $m=kn+z$).

The $\mathcal S_{a}^i$ dynamics only contains the first $n$ coordinates. $\mathcal S_{u}^i$ dynamics (containing the rest of coordinates) is used to obtain $\mathcal S^{i+1}$. Hence, $\mathcal S^{i}=\{\mathcal S^{i}_a, \mathcal S^{i+1}\}$ holds. Recursively,  the $\mathcal S^{i}$ dynamics is written as
\begin{align*}
\mathcal S_{a}^i: &~\bm D_{aa}^{(i)}\bm q^{(i)}_a + \bm D_{au}^{(i)}\bm q^{(i)}_u + \bm H^{(i)}_a=\bm B^{(i)}_a \bm u,\\
\mathcal S_{u}^i &=\left\{\mathcal S_{a}^{i+1},...,\mathcal S_{a}^k,\mathcal S^{k+1}\right\},
\end{align*}
where $\bm q^{(i)}_u$ is composed by $\bm q^{(i+1)}_a,\cdots \bm q^{(k)}_a, \bm q^{(k+1)}$. The coupling in $\mathcal S^{k+1}$ and $\mathcal S_{a}^k$ shows up only in $\bm q^{(k+1)}$ virtually. The original system $\mathcal S$ then can be rewritten into a series of cascaded subsystems as
\begin{equation}\label{Eq_CEIC}
\mathcal{S} \equiv\{\mathcal{S}_{a}^0, \underbrace{\mathcal{S}_{a}^1,\underbrace{ \mathcal{S}_{a}^2, \ldots, \underbrace{\mathcal{S}_{a}^k, \mathcal{S}^{k+1}}_{\mathcal{S}^k}\}}_{\mathcal S^2}}_{\mathcal{S}^1}
\end{equation}
where $\mathcal{S}_{a}^{k+1}=\mathcal S^{k+1}=\mathcal S_{u}^k$.

The BEM can still be used to characterize the balance target profile of each sub-order underactuated system.  Given the control input $\bm u$, the BEM for the underactuated system $\mathcal S^i$ is obtained by using its unaccentuated subsystem. The BEM $\mathcal{E}_i$ is defined
\begin{equation}\label{Eq_BEM_Si}
\mathcal{E}_{i}=\left\{\bm q_u^{(i+1),e}:  \bm \Gamma_{i+1}\left(\bm q_a^{(i+1)};\bm u\right)=\bm 0, \dot{\bm q}_a^{(i+1)},\ddot{\bm q}_a^{(i+1)}=\bm 0\right\},
\end{equation}
where $\bm \Gamma_{i+1}$ is obtained by using the dynamics of $\mathcal S^{i+1}_a$
\begin{align*}
\bm \Gamma_{i+1} &=\bm D^{(i+1)}_{aa}\ddot{\bm q}^{(i+1)}_a + \bm D^{(i+1)}_{au}\ddot{\bm q}^{(i+1)}_u + \bm H^{(i+1)}_a -\bm B_a^{(i+1)}\bm u.
\end{align*}
Clearly, $\mathcal{E}_i$ follows the BEM definition but only accounts for $\bm q_a^{(i+1)}$ (i.e., the $n-m$ coordinates in $\bm q_u^{(i)}$). While the rest of the unactuated coordinates $\bm q_u^{(i)}$ is untouched.

\begin{figure*}
  \centering
  \subfigure[]{
		\label{Fig_CEIC}
		\includegraphics[height=4cm]{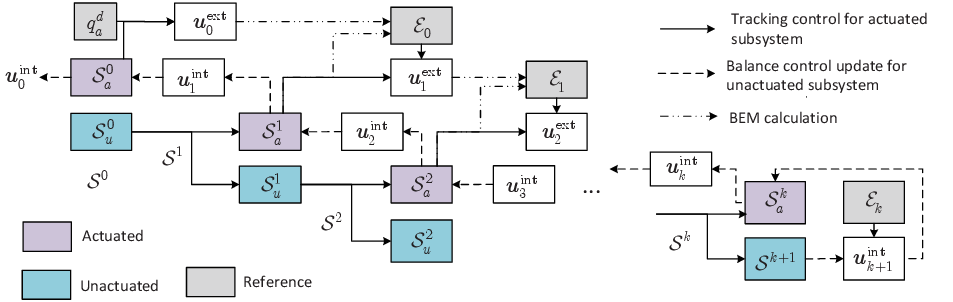}}
\subfigure[]{
		\label{Fig_EIC}
		\includegraphics[height=3.cm]{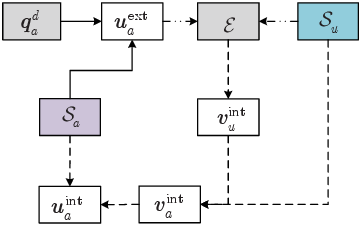}}
  \caption{Illustrative diagram of control design for $\mathcal S$. (a) CEIC-based control design. (b) EIC-based control design.}
  \label{Fig_Diagram_EIC}
\end{figure*}

\section{Cascaded Tracking Control Design}
\label{Sec_Control}

Based on the CEIC form, in this section we design the control input and show the stability of the closed-loop dynamics.

\subsection{Control Design}
Starting from $\mathcal S_{a}^0$ (the actuated subsystem of $\mathcal S$), we sequentially design the control input and obtain the corresponding BEM. The control input to drive $\bm q^{(0)}_a\rightarrow \bm q^{(0),d}_a$ can be designed using the feedback linearization technique as
\begin{equation}
  \bm u_0^\mathrm{ext} =\left(\bm B_a^{(0)}\right)^{-1}\left(\bm D^{(0)}_{aa}\bm v_0^\mathrm{ext} + \bm D^{(0)}_{au}\ddot{\bm q}^{(0)}_u + \bm H^{(0)}_a\right),
\end{equation}
where  $\bm v_0^\mathrm{ext} = \ddot{\bm q}^{(0),d}_a - \bm a_{0}\bm e_0- \bm b_{0}\dot{\bm e}_0$, $\dot{\bm e}_0 =\bm q^{(0)}_a- \bm q^{(0),d}_a$ is the tracking error, and $\bm a_{0}, \bm b_{0}$ are control gains. The design of $\bm u_0^\mathrm{ext}$ follows the same idea as shown in~\eqref{Eq_Actuated_Control} regardless of the numbers of unactuated coordinates.

Now let's consider the general case. If the control input for $\mathcal S^{i}$ is known, denoted as $\bm u_i^\mathrm{ext}$, we need to design the control for $\mathcal S^{i+1}$.  Within CIEC form, the connection between $\mathcal S^{i}$ and $\mathcal S^{i+1}$ is the dynamics of the first $n$ unactuated coordinates in $\mathcal S_{u}^i$. Therefore,  we only concern the first $n$ unactuated coordinates in $\bm q^{(i)}_u$ (i.e., $\bm q^{(i+1)}_a$, which appears in $\mathcal S^{i+1}_a$).

Obtaining the BEM using $\bm q^{(i+1)}_a$ is equivalently to inverting the $\mathcal S^{i+1}_{a}$ dynamics under the control design $\bm u=\bm u_i^\mathrm{ext}$ and the condition $\dot{\bm q}_a^{(i+1)}=\ddot{\bm q}_a^{(i+1)}=\bm 0$. Mathematically, we obtain BEM $\mathcal E_{i}$ by solving the implicit equation $\bm \Gamma_{i+1}$ in~\eqref{Eq_BEM_Si}. We denote the BEM solution as $\bm q^{(i+1),e}_a$, which becomes the reference trajectory for $\mathcal S^{i+1}$. The control input then can be updated  to enforce $\bm q^{(i+1)}_a \rightarrow \bm q^{(i+1),e}_a$. We design the $\bm u_{i+1}^\mathrm{ext}$
\begin{equation}\label{Eq_U_i+1}
\bm u_{i+1}^\mathrm{ext} =\left(\bm B_a^{(i+1)}\right)^{-1}\left(\bm D^{(i+1)}_{aa}\bm v_{i+1}^\mathrm{ext} + \bar{\bm H}^{(i+1)}_a\right)
\end{equation}
where $\bar{\bm H}^{(i+1)}_a= \bm D^{(i+1)}_{au}\ddot{\bm q}^{(i+1)}_u + \bm H^{(i+1)}_a$, $\bm v_{i+1}^\mathrm{ext} = \ddot{\bm q}^{(i+1),e}_a - \bm a_{i+1}\bm e_{i+1}- \bm b_{i+1}\dot{\bm e}_{i+1}$ is the auxiliary control and $\bm a_{i+1}, \bm b_{i+1} \in \mathbb{R}^{ n \times n}$.  The tracking error is defined as $\bm e_{i+1}=\ddot{\bm q}^{(i+1)}_a-\ddot{\bm q}^{(i+1),e}_a$.

Using the control input $\bm u_{i+1}^\mathrm{ext}$, we can solve the BEM $\mathcal E_{i+1}$ for $\mathcal S^{i+1}$ and design the control input $\bm u_{i+2}^\mathrm{ext}$. Recursively, we are able to obtain the control design for $\mathcal S^{k+1}$. The control input for $\mathcal S^{k+1}$ (the last subsystem) is
\begin{align*}
 \bm u_{k+1}^\mathrm{int}&=\left(\bm B^{(k+1)}\right)^+\left(\bm D^{(k+1)}\bm v_{k+1}^\mathrm{int}+ \bm H^{(k+1)}\right),\\
 \bm v_{k+1}^\mathrm{int} &= \ddot{\bm q}^{(k+1),e} - \bm a_{k+1}\bm e_{k+1}- \bm b_{k+1}\dot{\bm e}_{k+1},
\end{align*}
where $\bm a_{i+1}, \bm b_{i+1} \in \mathbb{R}^{z \times z}$. $\bm v_{k+1}^\mathrm{int}$ is the auxiliary control design that drives $\bm q^{(k+1)}$ to $\bm q^{(k+1),e}$.

The ultimate internal state of the original system becomes $\bm q^{(k+1)}$ and $\mathcal S^{k}$ is the simplest sub-order underactuated system with the property $\dim(\bm q_a^{(k)})\ge \dim(\bm q_u^{(k)})$. Given the balance control $\bm u_{k+1}^\mathrm{int}$, incorporating the balance control of $\bm q^{(k+1)}$ can be achieved by the EIC-based controller. We redesign the control input so that the virtually "actuated" coordinates ($\bm q_a^{(k)}$) will drive $\bm q_u^{(k)}$ to $\mathcal E_k$. Inserting $\bm u_{k+1}^\mathrm{int}$ and $\bm v_{k+1}^\mathrm{int}$ into $\mathcal S_{a}^k$ dynamics leads to
\begin{equation}\label{Eq_Sk_uint}
  \bm D_{aa}^{(k)} \ddot{\bm q}_a^{(k)}+\bm D_{au}^{(k)} \bm v_{k+1}^\mathrm{int}+\bm H_a^{(k)}=\bm B_a^{(k)} \bm u^\mathrm{int}_{k+1}.
\end{equation}
Clearly, in order to achieve $\bm q_u^{(k)}=\bm v_{k+1}^\mathrm{int}$, we need to revise $\bm q_a^{(k)}$ dynamics (i.e., $\mathcal S_{a}^k$), which is realized by redesigning the control input
\begin{subequations}\label{Eq_Uint_Sk}
\begin{align}
\bm u_k^\mathrm{int} & =\left(\bm B_a^{(k)}\right)^{-1}\left(\bm D_{aa}^{(k)} \bm v_k^\mathrm{int}+\bm D_{au,k+1}^{(k)} \bm q^{(i)}_u+\bm H_a^{(k)}\right), \\
\bm v_k^\mathrm{int} & =\left(\bm D_{aa}^{(k)}\right)^{-1}\left(\bm B_a^{(k)} \bm u_{k+1}^\mathrm{int}- \bm D_{au,k}^{(k)}\bm v_{k+1}^\mathrm{int} -\bm H_a^{(k)}\right).\label{Eq_Uint_Sk_b}
\end{align}
\end{subequations}
It is easy to verify $\bm q_u^{(k)}=\bm v_{k+1}^\mathrm{int}$ by replacing the controls~\eqref{Eq_Sk_uint} with those in~\eqref{Eq_Uint_Sk}. The control updating for $\bm u_k^\mathrm{int}$ follows a similar idea in~\eqref{Eq_Va_Int}. Under $\bm u_k^\mathrm{int}$, the balance of $\bm q^{(k)}$ is guaranteed.

For $\mathcal S^i$, $\bm u_i^{\mathrm{int}}$ is obtained by replacing $k$ with $i$ in~\eqref{Eq_Uint_Sk}. In particular, the $\bm v_i^{\mathrm{int}}$ is designed to update the virtually "actuated" coordinate $\bm q^{(i+1)}_a$ dynamics so that it drives $\bm q^{(i+1)}_a$ to BEM. The control $\bm  v_i^{\mathrm{int}}$ is
\begin{align}
\bm v_i^{\mathrm{int}}=&\left(\bm D_{a a}^{(i)}\right)^{-1}
\left(\bm B_a^{(i)} \bm u_{i+1}^{\mathrm{int}}-\bm D_{a u, i+1}^{(i)} \bm v_{i+1}^{\mathrm{int}} \right.\nonumber\\
&\left.\qquad-\sum\nolimits_{j=i+1}^k \bm D_{a u, j+1}^{(i)} \ddot{\bm q}_a^{(j+1)}-\bm H_a^{(i)}\right).
\end{align}
We only consider $\bm v_{i+1}^\mathrm{int}$ (the first $n-m$ unactuated coordinates of $\mathcal S^i$) in updating the motion of virtually actuated coordinates. We denote the final control as $\bm u_0^\mathrm{int}$. The diagram in Fig.~\ref{Fig_Diagram_EIC} shows the structure of the proposed control design. We sequentially decompose the system $\mathcal S^i$ and design control for actuated subsystem. When updating the control input, the $\mathcal S^{i+1}$ dynamics is recognized as the internal subsystem of $\mathcal S^i$ as shown in Fig.~\ref{Fig_CEIC}. However, in EIC-based control, the BEM is solved at once and the updated control needs to take of all unactuated coordinates (see Fig.~\ref{Fig_EIC}).

\subsection{Stability Analysis}

Firstly, we show that all coordinates of $\mathcal S^i$ under the control design $\bm u_i^\mathrm{int}$ are under active control. Secondly, the convergence of the tracking error for $\mathcal S^i$ is proved.

\begin{lem}\label{lemma_CEIC_dyna}
Given the highly underactuated balance system $\mathcal S$,  if $\mathcal S$ can be written into the CEIC form~\eqref{Eq_CEIC}, under the control input $\bm u_i^\mathrm{int}$, the closed-loop dynamics of $\mathcal S^i$  becomes
\begin{align*}
  \ddot{\bm q}^{(j)}_{a}&=\bm v_j^\mathrm{int},~   i\le j \le k,\\
  \ddot{\bm q}^{(k+1)}&=\bm v_{k+1}^\mathrm{int}.
  \end{align*}
\end{lem}

\begin{proof}
The proof can be found in Appendix~\ref{proof_lemma_CEIC}.
\end{proof}

The primary concern when applying EIC-based control to $\mathcal S$ is that certain coordinates would not display desired dynamics as shown in~\eqref{Eq_Sa_Ua_Int}. The result in Lemma~\ref{lemma_CEIC_dyna} indicates that each sub-order underactuated system is under active control design. Meanwhile, the constant input matrix assumption is no longer needed.

Next, we show that  $\bm q$ converge to $\{\mathcal{E}_i,...,\mathcal{E}_{k+1}\}$ ($\bm q_a^d$ can be viewed as $\mathcal E_0$). Based on the results in Lemma~\ref{lemma_CEIC_dyna}, the closed-loop dynamics of $\mathcal S^{k+1}$ under the control design $\bm u_i^\mathrm{int}$ becomes
\begin{align*}
  \ddot{\bm q}^{(k+1)}=\bm v_{k+1}^\mathrm{int}=\ddot{\bm q}^{(k+1),e} - \bm a_{k+1}\bm e_{k+1}- \bm b_{k+1}\dot{\bm e}_{k+1}.
\end{align*}
The $\mathcal S^{k+1}$ dynamics clearly is exponentially stable, if $\bm a_{k+1}$ and $\bm b_{k+1}$ are selected properly.

The preliminary control design $\bm u_i^\mathrm{ext}$ is used to obtain $\mathcal{E}_i$. $\bm \Gamma_{i+1}=\bm 0$ can be explicitly written as
\begin{equation}\label{Eq_BEM3}
\bm D_{au}^{(i + 1)}\ddot{\bm q}_u^{(i + 1)} + \bm H_a^{(i + 1)} - \bm B_a^{(i + 1)}\bm u_i^\mathrm{ext}=\bm 0
\end{equation}
under $\bm q^{(i+1)}_a=\bm q^{(i+1),e}_a$ and $\ddot{\bm q}^{(i+1)}_a=\ddot{\bm q}^{(i+1)}_a=\bm 0$. The above relationship~\eqref{Eq_BEM3} shall play a significant role in showing the convergence of $\bm q_a^{(i)}$. The control input $\bm u_{i+1}^\mathrm{int}$ is used to update $\bm u_{i}^\mathrm{int}$. We rewrite $\bm u_{i+1}^\mathrm{int}$ around $\bm q^{(i+1)}_a=\bm q^{(i+1),e}_a, \ddot{\bm q}^{(i+1)}_a=\ddot{\bm q}^{(i+1)}_a=\bm 0$,
\begin{align}\label{Eq_u_ext_appro}
\bm u_{i+1}^\mathrm{int}
& =\left(\bm B_a^{(i+1)}\right)^{-1}\left(\bm D_{a u}^{(i+1)} \ddot{\bm q}_u^{(i+1)}+\bm H_a^{(i+1)}\right)\big\vert_{\bm x_q^{(i+1),e}}+\bm o_i\nonumber \\
& = \left(\bm B_a^{(i+1)} \right)^{-1}\bm B_a^{(i+1)}\bm u_i^\mathrm{ext} \big\vert_{\bm x_q^{(i+1),e}}  + \bm o_i \nonumber \\
& =\bm u_i^\mathrm{ext}+\bm o_i.
\end{align}
where~\eqref{Eq_BEM3} is used to simplify the above equation. $\bm x_q^{(i+1),e}=\{\bm q^{(i+1),e}_a~\bm 0,\bm 0\}$ and $\bm o_i$ denotes perturbations including the higher order term and $\left(\bm B_a^{(i+1)}\right)^{-1} \bm D_{a a}^{(i+1)} \bm v_{i+1}^{\mathrm{ext}}$.

To proceed, substituting~\eqref{Eq_u_ext_appro} into $\ddot{\bm q}^{(i)}_{a}=\bm v_i^\mathrm{int}$ and using Lemma~\ref{lemma_CEIC_dyna} yields $\ddot{\bm q}_a^{(i)}=\bm v_i^\mathrm{int}=\bm v_i^\mathrm{ext}+\bm O_i$, where $\bm O_i=\left(\bm{D}_{a a}^{(i)}\right)^{-1} \bm{B}_a^{(i)} \bm{o}_i$. The closed-loop dynamics becomes
\begin{subequations}
\begin{align}
\ddot {\bm e}_{i} & =-\bm a_i\bm e_{i}-\bm b_i\bm e_{i}+ \bm O_{i},  i  \le k \\
\ddot {\bm e}_{k+1} & =-\bm a_{k+1}\bm e_{k+1}-\bm b_{k+1}\bm e_{k+1}.
\end{align}
\end{subequations}
Let $\bm \xi= [\bm e_0^T~\dot{\bm e}_0^T~...~\bm e_{k+1}^T~\dot{\bm e}_{k+1}^T]^T$ be the error vector. We rewrite the error dynamics into the following compact form
\begin{align}
\mathcal{S}_{e}:  \dot{\bm \xi}&=
\begin{bmatrix}
    \bm 0 & \bm I &\cdots & \bm 0 & \bm 0\\
    -\bm a_0 & -\bm b_0 &\cdots & \bm 0 & \bm 0\\
    & & \ddots & &    \\
    \bm 0 & \bm 0 &\cdots & \bm 0 & \bm I\\
    \bm 0 & \bm 0 &\cdots & -\bm a_{k+1} &-\bm b_{k+1}
  \end{bmatrix}\bm \xi
  + \begin{bmatrix}
  \bm 0 \\
  \bm O_0 \\
  \vdots \\
  \bm 0 \\
  \bm 0
  \end{bmatrix}\nonumber\\
  &\triangleq \bm A\bm \xi + \bm O_{\xi}.
\end{align}
If the gains $\{\bm a_j, \bm b_j\}, j=i,...,k+1$ are properly selected such that $\bm A$ is Hurwitz, $\bm \xi$ can be shown converging to zero under perturbations. Assume that the perturbation term is affine to tracking errors as $\norm{\bm O_\xi}\le c_1\norm{\bm \xi} +c_2$ for $c_1$ and $c_2>0$.

We take the Lyapunov function candidate
$V=\frac{1}{2}\bm \xi^T \bm \xi$. It is easy to show that
\begin{align*}
 \dot{V}&=\bm \xi^T \bm{A} \bm \xi+\bm \xi^T \bm{O}_{\bm \xi}  \leq \lambda_1(\bm{A})\norm{\bm \xi}^2+\norm{\bm \xi}\left(c_1\norm{\bm \xi} +c_2\right) \\
& =\left[\lambda_1(\bm{A})+c_1\right]\norm{\bm \xi}^2+c_2\norm{\bm \xi}
\end{align*}
where $\lambda_1(\bm{A})$ denotes the greatest eigenvalue of $\bm A$. If $\lambda_1(\bm{A})+c_1<0$, the tracking error is exponentially decreasing under perturbation.

The control design is based on the CIEC form and thus the system dynamics should satisfy certain conditions. Here we summarize the conditions:
\begin{itemize}
  \item fully ranked matrix for each sub-order underactuated system $\rank(\bm D^{(i)}_{aa})=\rank(\bm D^{(i)}_{au})=\rank(\bm B^{(i)}_{a})=n$, $i\le k$ and $\rank(\bm D^{(k+1)}_{aa})=\rank(\bm D^{(k+1)}_{au})=\rank(\bm B^{(k+1)}_{a})=z$;
  \item $\bm D_{a a}^{(i+1)}-\bm B_a^{(i+1)}\left(\bm B_a^{(i)}\right)^{-1} \bm D_{a u, i+1}^{(i)} \neq \bm 0$ to guarantee that the each actuated subsystem can display the designed dynamics.
\end{itemize}

\begin{figure}[h]
    \centering
    \includegraphics[height=4.5cm]{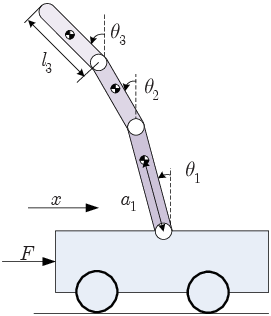}
    \caption{Schematics of a triple-inverted pendulum on a cart. The three joints $\theta_1$, $\theta_2$, and $\theta_3$ are unactuated.}
    \label{Fig_Cart_Triple}
\end{figure}
\section{Results}
\label{Sec_Result}

\begin{figure*}
  \centering
  \subfigure[]{
		\label{Fig_Triple_1}
		\includegraphics[height=4.8cm]{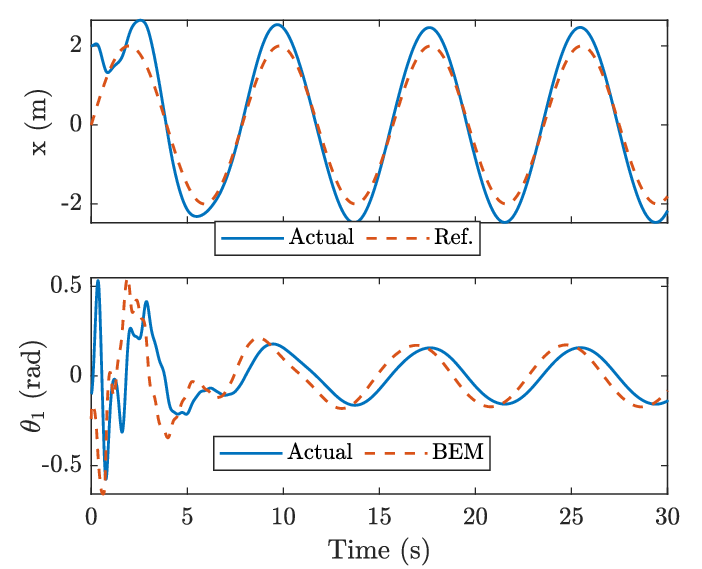}}
\hspace{-4.5mm}
\subfigure[]{
		\label{Fig_Triple_2}
		\includegraphics[height=4.8cm]{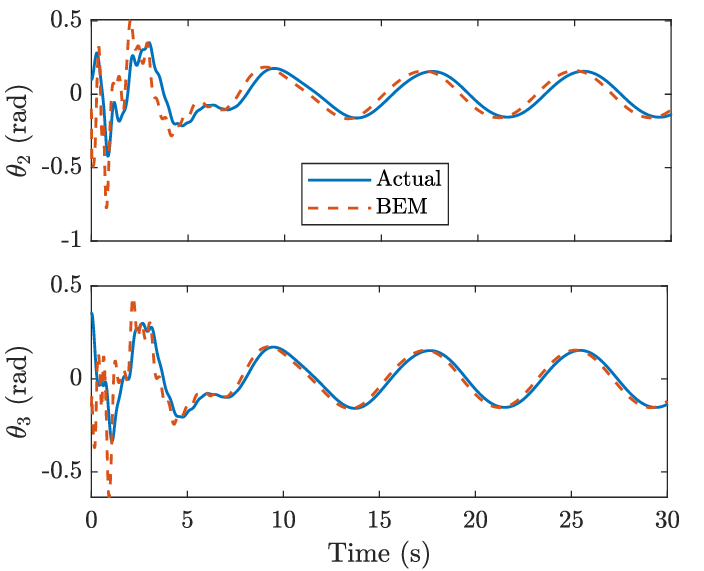}}
\hspace{-4.5mm}
\subfigure[]{
		\label{Fig_Triple_error1}
		\includegraphics[height=4.8cm]{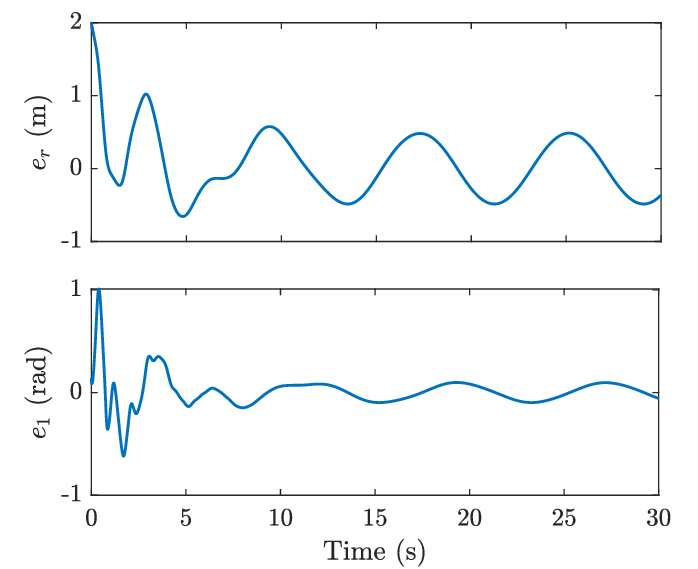}}
\subfigure[]{
		\label{Fig_Triple_error2}
		\includegraphics[height=4.8cm]{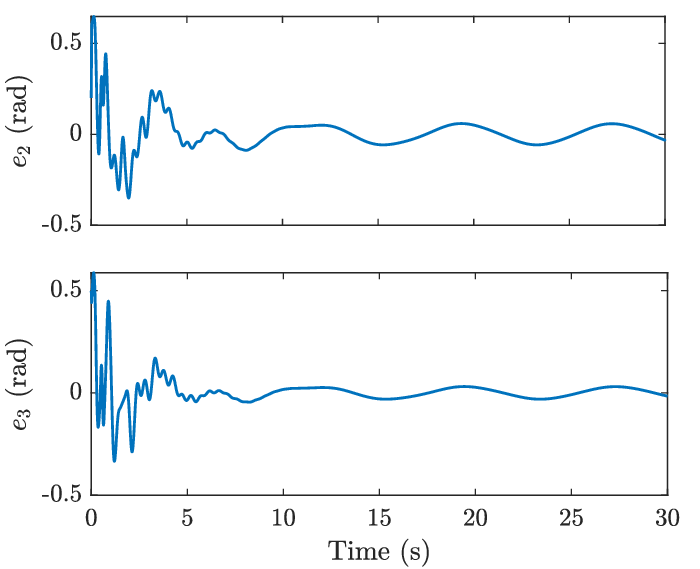}}
\hspace{-4.5mm}
\subfigure[]{
		\label{Fig_Triple_EIC1}
		\includegraphics[height=4.8cm]{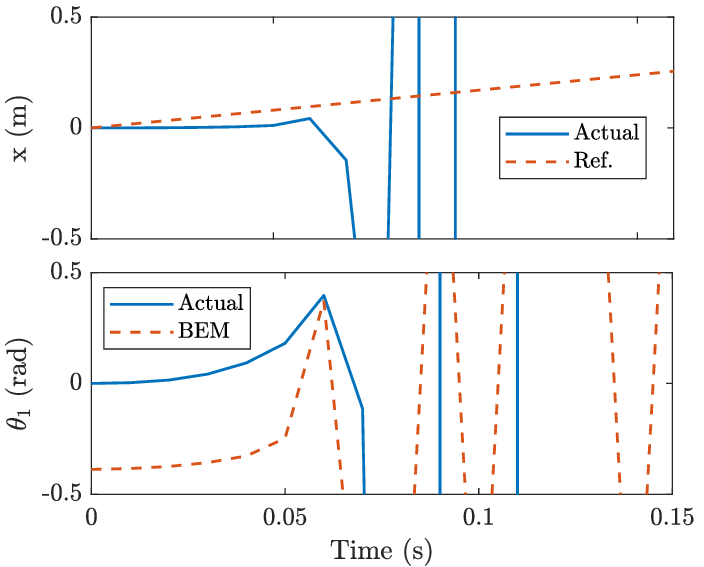}}
\hspace{-4.5mm}
\subfigure[]{
		\label{Fig_Triple_EIC2}
		\includegraphics[height=4.8cm]{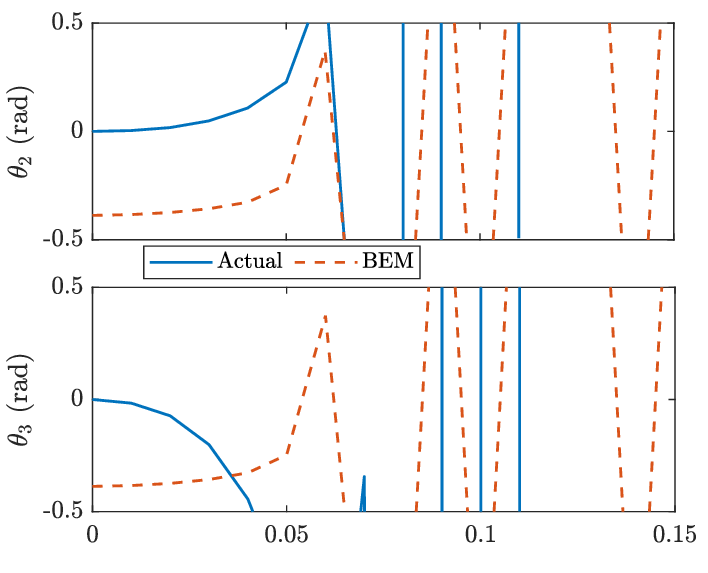}}
  \caption{Tracking control of a triple-inverted pendulum cart. (a) and (b) shows the cart position and pendulum angles under the proposed control. (c) and (d) shows the tracking errors. (e) and (f) shows the cart position and pendulum rotation angles under the EIC-based control.}
  \label{Fig_Triple_control}
\end{figure*}
We present the simulation result to demonstrate and validate the proposed control design in this section. Fig.~\ref{Fig_Cart_Triple} shows a triple-inverted pendulum system on a moving cart. With four DOFs, only the cart is actuated and moves left/right to follow the given reference trajectory while keeping the inverted pendulum balanced around the vertical position. The dynamics model of a cart-triple inverted pendulum system~\cite{Medrano1997} can be written into the form of $\mathcal S$ with
\begin{align*}
\bm D&=\begin{bmatrix}
M_t & -M_1\c_1 & -M_2\c_2 & -M_3\c_3 \\
-M_1 \c_1 & I_1 & M_2 l_1 \c_{21} &M_3 l_1 \c_{31} \\
-M_2 \c_2 & M_2 l_1 \c_{21} & I_2 & M_3 l_2 \c_{32} \\
-M_3 \c_3 & M_3 l_1 \c_{31} & M_3 l_2 \c_{32} & I_3
\end{bmatrix},\\
\bm C&=\begin{bmatrix}
0 & M_1 \dot{\theta}_1 \s_1 & M_2 \dot{\theta}_2 \s_2 & M_3 \dot{\theta}_3 \s_3 \\
0 & 0 & -M_2 l_1 \dot{\theta}_2 \s_{21} & -M_3 l_1 \dot{\theta}_3 \s_{31} \\
0 & M_2 l_1 \dot{\theta}_1 \s_{31}& 0 & -M_3 l_2 \dot{\theta}_3 \s_{32} \\
0 & M_3 l_1 \dot{\theta}_1 \s_{31} & M_3 l_2 \dot{\theta}_2 \s_{32} & 0
\end{bmatrix},\\
\bm G&=\begin{bmatrix}
0 \\
-M_1 g \s_1 \\
-M_2 g \s_2\\
-M_3 g \s_3
\end{bmatrix},~
\bm B=\begin{bmatrix}
1\\
0\\
0\\
0
\end{bmatrix},
\end{align*}
where $\s_i=\sin\theta_i$, $\c_i=\cos\theta_i$, $\s_{ij}=\sin(\theta_i-\theta_j)$, and  $\c_{ij}=\cos(\theta_i-\theta_j)$. The variables are defined as $ M_t=m_c+m_1+m_2+m_3$, $M_1=m_1 a_1+(m_2+m_3) l_1$, $M_2=m_2 a_2+m_3 l_2$, $M_3=m_3 a_3$, $I_1 =J_1+m_1 a_1^2+(m_2+m_3) l_1^2$, $I_2 =J_2+m_2 a_2^2+m_3 l_2^2$, $I_3 =J_3+m_3 a_3^2$. The length and distance from the joint to COM of each link are $l_i$ and $a_i$ respectively. The mass and the moment of inertia of each link are $m_i$ and $J_i$. The gravity constant is $g$.

Let $q_a^{(1)}=x, q_a^{(2)}=\theta_1, q_a^{(3)}=\theta_2, q^{(4)}=\theta_3$, we rewrite the system dynamics into the CIEC form. In particular, the $\mathcal S^1_a$ dynamics is explicitly given as
\begin{align*}
\mathcal S_{a}^1:& ~\left(J_1 M_t-M_1^2 \c_1^2\right) \ddot{\theta}_1+M_2\left(\c_{21} l_1 M_t-\c_2 \c_1 M_1\right) \ddot{\theta}_2 \\
& +M_3\left(\c_{32} l_2 M_t-\c_3 \c_1 M_1\right) \ddot{\theta}_3 \\
&-\left(M_2 l_1 \dot{\theta}_2^2 \s_{12}+M_3 l_1 \dot{\theta}_3^2 \s_{32}+M_1 g \s_1\right) M_t  \\
& +M_1 \c_1\left(M_1 \dot{\theta}_1^2 \s_1+M_2 \dot{\theta}_2^2 \s_2-M_3 \dot{\theta}_3^2 \s_3\right)=M_1 \c_1 u
\end{align*}
where the cart acceleration does not show up as designed. The moment of inertia $J_1 M_t-M_1^2 \c_1^2$ and the input matrix $M_1 \c_1$ can be shown away from 0 for appropriate trajectory. Then the inverse of those matrixes exists. $\mathcal S_{a}^2$ and $\mathcal S^3$ can be obtained accordingly.

The reference trajectory of the cart is $x^d=2\sin(0.8t)$. The control gains are $a_0=0.8, b_0=2.5, a_1=35, b_1=3.5, a_2=38, b_2=4.85, a_3=50, b_3=15$. The initial position of the system is $x= 2$~m, $\theta_1=-0.1$~rad, $\theta_2=0.1$~rad, $\theta_3=0.35$~rad, which is far away from the static equilibrium. Fig.~\ref{Fig_Triple_control} shows the simulation results under EIC-based control and the proposed control design. Under the CIEC-based control design, the cart follows the given reference trajectory, and all three unactuated links were kept balanced on the BEM as shown in Fig.~\ref{Fig_Triple_1} and Fig.~\ref{Fig_Triple_2}. While the system becomes unstable (see Fig.~\ref{Fig_Triple_EIC1} and Fig.~\ref{Fig_Triple_EIC2}) when the EIC-based control is applied, which validates the analysis in Section~\ref{Sec_Model}. In EIC-based control, the cart position coordinates carry the task of balancing all three links. While the CIEC-based control only assigns the task of balance link $\theta_1$ to the cart.

The tracking errors are shown in Fig.~\ref{Fig_Triple_error1} and Fig.~\ref{Fig_Triple_error2}.  We further summarize the steady tracking error in Table.~\ref{Table_error} (mean and standard deviation). The relative error is obtained by normalizing the tracking error with the reference' (or BEM profile) amplitude. Since the system is in a cascaded form, the tracking error in the internal system would affect the tracking performance in the external system. It is observed in Table~\ref{Table_error} that $|e_1|>|e_2|>|e_3|$ in terms of the mean errors for both absolute and relative errors. The part of $\bm q_a^{(i)}$ motion effect serves as the control input to drive $\bm q_a^{(i-1)}$ to its BEM. $q_a^{(3)}$ (equivalently $\theta_2$) would not achieve the best tracking performance until the $q_a^{(4)}$ (equivalently $\theta_3$) perfectly follows its BEM. In such a way, the task of balancing $\mathcal S^{k+1}$ is placed at the highest priority and all other unactuated systems are balanced one by one sequentially.

\renewcommand{\arraystretch}{1.3}
\setlength{\tabcolsep}{0.045in}
\begin{table}
  \centering
    \caption{Tracking errors of the triple-inverted pendulum}
    \label{Table_error}
  \begin{tabular}{|c|c|c|c|c|}
    \hline\hline
     & $|e_r|$ & $|e_1|$ & $|e_2|$ & $|e_3|$ \\
     \hline
    Absolute (rad) & $0.31\pm0.15$  & $0.06\pm 0.03$  & $0.04\pm 0.02$  & $0.02\pm 0.01$  \\
    \hline
    Relative ($\%$) & $15.6\pm7.4$  &$35.2\pm17.2$   & $22.7\pm11.1$  &  $12.4\pm 6.0$ \\
    \hline\hline
  \end{tabular}
\end{table}

\section{Conclusion}
\label{Sec_Conclusion}

This paper proposed a cascaded nonlinear control framework for highly underactuated balance robots (i.e., there are more unactuated coordinates). To achieve simultaneous trajectory tracking and balance control, the proposed framework converts a highly underactuated robot system to a series of cascaded virtually actuated subsystems. The tracking control inputs are sequentially designed layer by layer until the last subsystem. The control input then is updated from the last subsystem to the first one to incorporate the balance task. Under such a sequential design and updating, we show the closed-loop system dynamics is stable. We validate the control design with numerical simulation on a triple-inverted pendulum cart system. In the future, we plan to extend such a framework with machine learning-based techniques to achieve guaranteed performance and test the framework with physical robot systems.

\appendices

\section{Proof for Lemma~\ref{lemma_CEIC_dyna}}\label{proof_lemma_CEIC}
Inserting $\bm u_i^\mathrm{int}$ into $\mathcal S_{a}^i$  leads to
$\bm D_{a a}^{(i)} \ddot{\bm q}_a^{(i)}+\bm D_{a a}^{(i)} \ddot{\bm q}_u^{(i)}+\bm H_a^{(i)} = \bm B_a^{(i)}\left(\bm B_a^{(i)}\right)^{-1}\left(\bm D_{a a}^{(i)}\bm v_i^\mathrm{int}+\bm D_{a u}^{(i)} \ddot{\bm q}_u^{(i)}+\bm H_a^{(i)}\right)$. After simplification, we can obtain that $\ddot{\bm q}^{(i)}_{a}=\bm v_i^\mathrm{int}$.

Next we show that $\mathcal S^{i+1}_a$ under the control input $\bm u_i^\mathrm{int}$ displays the dynamics behavior $\ddot{\bm q}_a^{(i+1)}=\bm v_{i+1}^\mathrm{int}$. Substituting $\bm u_i^\mathrm{int}$ into $\mathcal  S^{i+1}_a$ we obtain
\begin{equation}\label{Eq_dif}
\bm D_{a a}^{(i+1)} \ddot{\bm q}_a^{(i+1)}+\bm D_{a u}^{(i+1)} \ddot{\bm q}_u^{(i+1)}+\bm H_a^{(i+1)} =\bm B_a^{(i+1)} \bm u_{i}^\mathrm{int}
\end{equation}
The right hand side of~\eqref{Eq_dif} is simplified by inserting the explicit form of $\bm u_{i}^\mathrm{int}$ as
\begin{align*}
\mathrm{RHS}
&=\bm B_a^{(i+1)}\left(\bm B_a^{(i)}\right)^{-1}\left(\bm D_{a a}^{(i)} \bm v_i^\mathrm {int}+\bm D_{a u}^{(i)} \ddot{\bm q}_u^{(i)}+\bm H_a^{(i)}\right),
\end{align*}
where
\begin{align*}
& \bm B_a^{(i+1)}\left(\bm B_a^{(i)}\right)^{-1} \bm D_{a a}^{(i)} \bm v_i^\mathrm{int} \\
& =\bm B_a^{(i+1)} \bm u_{i+1}^{\mathrm{int }}-\bm B_a^{(i+1)}\left(\bm B_a^{(i)}\right)^{-1}\left(\bm D_{a u}^{(i)}
\begin{bmatrix}
\bm v_{i+1}^{\mathrm {int }} \\
\ddot{\bm q}_u^{(i+1)}
\end{bmatrix}+\bm H_a^{(i)}\right) \\
& =\bm D_{a a}^{(i+1)} \bm v_{k+1}^\mathrm{int}+\bm D_{a u}^{(i+1)} \ddot{\bm q}_u^{(i+1)}+\bm H_a^{(i+1)}\\
&\qquad\qquad-\bm B_a^{(i+1)}\left(\bm B_a^{(i)}\right)^{-1}\left(\bm D_{a u}^{(i)}
\begin{bmatrix}
\bm v_{i+1}^\mathrm{int} \\
\ddot{\bm q}_u^{(i+1)}
\end{bmatrix}+\bm H_a^{(i)}\right).
\end{align*}
Thus, the right-hand side of~\eqref{Eq_dif} becomes
\begin{align*}
\mathrm{RHS} =&\bm D_{aa}^{(i + 1)}\bm v_{k + 1}^\mathrm{int} + \bm D_{au}^{(i + 1)}\ddot {\bm q}_u^{(i + 1)} + \bm H_a^{(i + 1)}\\
&- \bm B_a^{(i + 1)}{\left( {\bm B_a^{(i)}} \right)^{ - 1}}\bm D_{au}^{(i)}
\begin{bmatrix}
  \ddot{\bm q}_a^{(i + 1)} - \bm v_{i + 1}^\mathrm{int} \\
  \bm 0
\end{bmatrix}.
\end{align*}
Using above equation, \eqref{Eq_dif} is rewritten into
\begin{equation*}
\left[\bm D_{a a}^{(i+1)}-\bm B_a^{(i+1)}\left(\bm B_a^{(i)}\right)^{-1} \bm D_{a u, i+1}^{(i)}\right] \left(\ddot{\bm q}_a^{(i+1)}-\bm v_{i+1}^{\mathrm{int}}\right)=\bm 0.
\end{equation*}
If $\bm D_{a a}^{(i+1)}-\bm B_a^{(i+1)}\left(\bm B_a^{(i)}\right)^{-1} \bm D_{a u, i+1}^{(i)} \neq \bm 0$, the solution for above equation becomes $\ddot{\bm q}_a^{(i+1)}=\bm v_{i+1}^\mathrm{int}$, which is exactly the designed control input. The proof is continued until $\mathcal S^{k+1}$. Due to the page limit, it is not presented here.

\bibliographystyle{IEEEtran}
\bibliography{HanRef}
\end{document}